\documentclass[letterpaper]{article} % DO NOT CHANGE THIS
\usepackage{aaai25}  % DO NOT CHANGE THIS
\usepackage{times}  % DO NOT CHANGE THIS
\usepackage{helvet}  % DO NOT CHANGE THIS
\usepackage{courier}  % DO NOT CHANGE THIS
\usepackage[hyphens]{url}  % DO NOT CHANGE THIS
\usepackage{graphicx} % DO NOT CHANGE THIS
\usepackage{natbib}  % DO NOT CHANGE THIS AND DO NOT ADD ANY OPTIONS TO IT
\usepackage{caption} % DO NOT CHANGE THIS AND DO NOT ADD ANY OPTIONS TO IT
\usepackage{algorithm}
\usepackage{algorithmic}
\usepackage{enumitem}
\usepackage{amsmath}
\usepackage{amssymb}
\usepackage{multirow}
\usepackage{xcolor}
\usepackage{colortbl}
\usepackage{stackengine}
\newcommand{\model}{NAS}
\usepackage{graphicx}
\usepackage{caption}
\usepackage{lipsum}
\usepackage{booktabs}
\usepackage{bm}

\usepackage{newfloat}
\usepackage{listings}
\DeclareCaptionStyle{ruled}{labelfont=normalfont,labelsep=colon,strut=off} % DO NOT CHANGE THIS
\floatstyle{ruled}
\newfloat{listing}{tb}{lst}{}
\floatname{listing}{Listing}
\title{Enhancing Fine-Grained Vision-Language Pretraining \\ with Negative Augmented Samples}
\author{
    Yeyuan Wang \textsuperscript{\rm 1}$^*$, 
    Dehong Gao \textsuperscript{\rm 2}\thanks{These authors contributed equally and $^\dagger$corresponding authors.}, 
    Lei Yi \textsuperscript{\rm 3}, 
    Linbo Jin \textsuperscript{\rm 3},
    Jinxia Zhang \textsuperscript{\rm 4}, \\
    Libin Yang \textsuperscript{\rm 2}$^\dagger$, 
    Xiaoyan Cai \textsuperscript{\rm 1}$^\dagger$
    } 
\affiliations{
    \textsuperscript{\rm 1}School of Automation, Northwestern Polytechnical University, Xi'an, Shaanxi, China \\
    \textsuperscript{\rm 2}School of Cybersecurity, Northwestern Polytechnical University, Xi'an, Shaanxi, China \\
    \textsuperscript{\rm 3}Alibaba Group, Hangzhou, Zhejiang, China \\
    \textsuperscript{\rm 4}School of Automation, Southeast University, Nanjing, Jiangsu, China \\
    \{wangyeyuan, dehong.gdh, libiny, xiaoyanc\}@nwpu.edu.cn \\
    \{yilei.yi, yuyi.jlb\}@alibaba-inc.com \\
    \{jinxiazhang\}@seu.edu.cn  \\
}

\usepackage{bibentry}
\begin{document}

\maketitle

\begin{figure*}[!htbp]
  \centering
  \includegraphics[width=\linewidth]{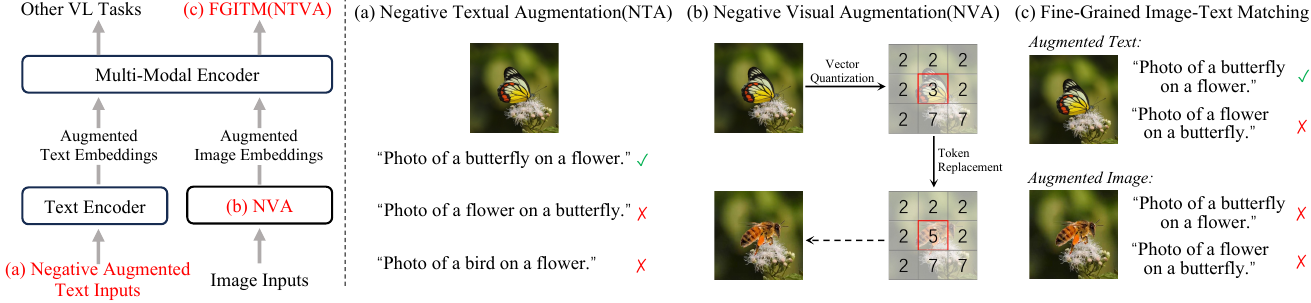}
  \caption{Fine-grained enhanced VLP architecture. NTA constructs the hard negative text samples for the language modality(a); We discrete the visual representation and construct the hard negative image samples for the visual modality(b); FGITM is proposed to leverage the fine-grained negative image and text samples to enhance the fine-grained capability(c).}
  \label{fig:teaser}
\end{figure*}

\begin{abstract}
Existing Vision-Language Pretraining (VLP) methods have achieved remarkable improvements across a variety of vision-language tasks, confirming their effectiveness in capturing coarse-grained semantic correlations. 
However, their capability for fine-grained understanding, which is critical for many nuanced vision-language applications, remains limited. 
Prevailing VLP models often overlook the intricate distinctions in expressing different modal features and typically depend on the similarity of holistic features for cross-modal interactions. 
Moreover, these models directly align and integrate features from different modalities, focusing more on coarse-grained general representations, thus failing to capture the nuanced differences necessary for tasks demanding a more detailed perception.
In response to these limitations, we introduce Negative Augmented Samples(\model), a refined vision-language pretraining model that innovatively incorporates \model~ to specifically address the challenge of fine-grained understanding. 
\model~ utilizes a Visual Dictionary(VD) as a semantic bridge between visual and linguistic domains. 
Additionally, it employs a Negative Visual Augmentation(NVA) method based on the VD to generate challenging negative image samples.
These samples deviate from positive samples exclusively at the token level, thereby necessitating that the model discerns the subtle disparities between positive and negative samples with greater precision. 
Comprehensive experiments validate the efficacy of~\model~components and underscore its potential to enhance fine-grained vision-language comprehension.
\end{abstract}


\section{Introduction}

Multi-modal machine learning, which aims to process and relate information of multiple modalities, is a domain that has a significant impact on general artificial intelligence~\cite{multimodal}.
Among these modalities, there has been a surge of interest in vision-language multi-modal research, as the vision and language modalities are widely used and closely intertwined in human daily life~\cite{Wang22:CAAI}.
However, current vision-language pretraining models mainly focus on capturing the overall relationship between vision and language~\cite{OFA,Wang_CVPR23_BEiT3, ji2023seeing}, often overlooking the more nuanced, local interactions~\cite{fashionbert,wei2021fine}.
The ability of modeling the local relationship, which we refer to as fine-grained capability, is crucial in various artificial intelligence domains such as medicine~\cite{Fang_ACMM20_medical,WangFGCCA21_medical}, agriculture~\cite{Hou_ICCV17_agriculture, Van_CVPR18_argriculture}, and e-commerce~\cite{Pang_CVPR19_Product, bai2020products10k}.
Therefore, it is necessary to conduct in-depth research to better understand and model the fine-grained attributes of both vision and language modalities.

Achieving this goal relies on two pivotal advancements: \textbf{fine-grained feature extraction} (the accurate extraction of subtle information from each input modality), and \textbf{fine-grained modality alignment}(the precise calibration of multi-modal features).
%extraction
According to \textbf{fine-grained feature extraction}, the discrete tokens are frequently selected as the fine-grained text features~\cite{bert} for language modality, while various image features (from the single pixels to patches or region features) are selected for visual modality~\cite{huang2020pixelbert, chen2020uniter, vilt, zeng2022multi}.
For example, salient visual regions can be located with pretrained object detectors as visual region features~\cite{li2020oscar, cho2021unifying, hu2022scaling}. 
The kaleidoscope-like patches are leveraged to represent multi-scale visual features~\cite{zhuge2021kaleidobert}.
These image features have propelled the advancement of VLP in the general domain; however, researchers still struggle to achieve fine-grained capability due to the huge semantic gaps between those \textbf{discrete} language tokens and these \textbf{continuous} visual features~\cite{zhao2023mamo}.
%alignment
According to \textbf{fine-grained modality alignment}, researchers apply contrastive learning, such as CLIP~\cite{clip}, to align images and sentences globally. 
The later extensions explored patch-token interactive methods to capture the fine-grained correlation~\cite{filip}.
Recently, researchers attempted to improve the fine-grained capability through Negative Textual Augmentation(NTA), which has made significant progress~\cite{negclip,zhang2023contrasting,huang2023structureclip}.
As shown in Figure~\ref{fig:teaser} (a), these NTA approaches employ either auxiliary models or syntax-based algorithms to generate negative text samples deliberately misaligned with corresponding images~\cite{negclip,doveh2023teaching,singh2023coarsetofine,huang2023structureclip}. 
Alongside the Image-Text Matching (ITM) task, these hard negative samples reinforce the VLP model to align the language and visual modalities fine-grainedly.
Although the construction of hard negative samples for language modality is well-documented due to the sparsity of the text space and advancements in Large Language Models (LLMs)~\cite{Parcalabescu_2022,liu2023visual}, 
the construction for visual modality is hindered by the complexity of visual signals~\cite{peng2024synthesize}.

This paper addresses the issue of the Negative Textual and Visual Augmentation(NTVA) method, which creates hard negative samples for both language and visual modality.
To tackle these challenges, we propose an innovative multi-modal model, called \model.
As shown in Figure~\ref{fig:teaser} (b), we integrate a VD into VLP model, which is regarded as the semantic abstraction of visual raw features, to bridge the semantic gaps between modalities.
The VD effectively quantifies continuous visual input into discrete tokens, easing fine-grained extraction and improving generalizability.
Furthermore, we introduce a novel NVA approach, leveraging semantic-aware token replacement based on the VD to foster fine-grained alignment by constructing negative image samples.
As shown in Figure~\ref{fig:teaser} (b) and Figure~\ref{fig:teaser} (c), our \model~ creates negative image samples by altering tokens based on the global and local feature similarities of text and image inputs.
These samples provoke the VLP models to pay more attention to detail alignment through FGITM task.
To evaluate the effectiveness of our proposed model, we conducted experiments on downstream fine-grained multi-modal tasks.
The results demonstrate that our model outperforms existing VLP models significantly.
In summary, our \textbf{contributions} are threefold:
\begin{itemize}[leftmargin=12pt]
    \item We first propose the NTVA method to simultaneously construct hard negative textual and visual samples, which can significantly improve the fine-grained capability together with the FGITM task. The NTVA method is a general data construction method that can be applied in related image fine-grained tasks.
    \item We introduce a novel VLP model named \model, which applies the NTVA method to VLP models. Using the ALBEF structure as framework, \model~ significantly improves the fine-grained capability of VLP models.
    \item Through comprehensive experiments on the ARO, Winoground, and VALSE datasets, we substantiate the efficacy of \model. The results confirm that our proposed NTVA approach sets a new SOTA in these datasets.
\end{itemize}

\begin{figure*}[!htbp]
  \centering
  \includegraphics[width=\linewidth]{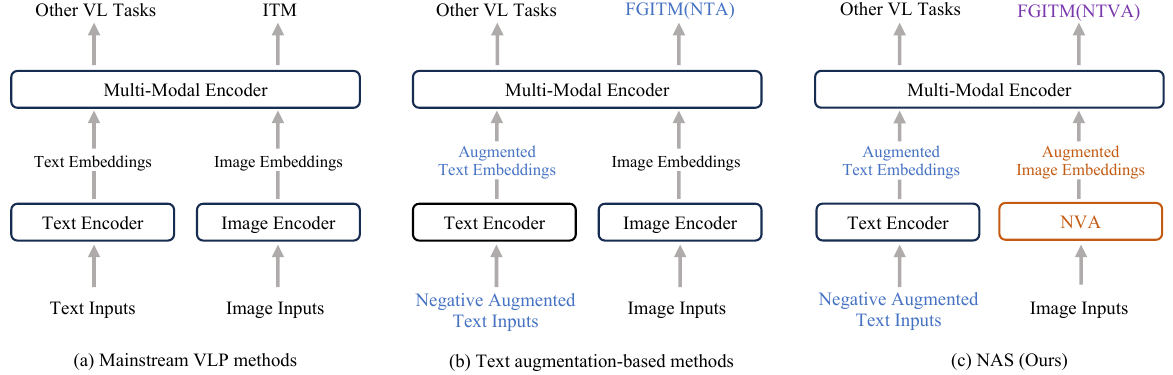}
  \caption{Comparison of our fine-grained \model~ to other VL frameworks. Mainstream VLP methods utilize two "Dual Tower" encoders and use a multi-modal encoder for deep fusion of multi-modal features(e.g., ALBEF~\cite{albef} and METER~\cite{dou2022empirical})(a), NTA-based methods construct augmented negative text samples to enhance VLP model's fine-grained ability with FGITM(e.g., VL-Match~\cite{bi2023vl} and ViLTA~\cite{wang2023vilta})(b), our \model~ introduces a NVA module to construct augmented negative image features, together with NTVA to enhance the VLP fine-grained capability with FGITM in an end-to-end manner(c).}
  \label{fig:comparison}
\end{figure*}

\section{Related Work}

\subsection{Visual Dictionary in Vision-Language}

Visual Dictionary is widely used in VLP for transforming continuous visual features into discrete, higher-level representations~\cite{soho, fdt, zheng2024iterated}. 
This transformation aids in alleviating semantic gap and capturing more nuanced visual semantic concepts.
% Given that the nearest neighbor search is non-differentiable, VD optimization typically involves a stop-gradient mechanism~\cite{chen2020exploring}.
Within the landscape of multi-modal research, the use of VD has begun to play a pivotal role~\cite{unimo2, fdt}.
SOHO~\cite{soho} leverages the VD to address semantic discrepancies.
Similarly, UNIMO2~\cite{unimo2} employs the VD as a cornerstone for modality alignment, effectively utilizing both uni-modal and multi-modal data streams.
The FDT framework~\cite{fdt} quantizes multi-modal features through a unified VD, further reinforcing the alignment across modalities.
Moreover, IL-CLIP~\cite{zheng2024iterated} introduces an iterated learning algorithm based on the unified VD, which improves compositionality in large vision-language models.

Our approach employs the VD to bridge the semantic gap between different modal features. 
We address mode collapse in VD learning by updating the dictionary with an exponential moving average mechanism, which improves both VD learning and training stability and setting the stage for implementing our proposed NVA method.

\subsection{Data Augmentation for Vision-Language}

Data augmentation (DA) is widely applied in computer vision and has expanded into the realm of VLP~\cite{mu2022slip, li2022supervision}.
Recent studies employ NTA to construct fine-grained hard negative sentences with similar structures but different semantics~\cite{negclip, huang2023structureclip, Momeni_2023_ICCV}.
In the visual domain, Syn-CLIP~\cite{cascante2023going} exploits 3D simulation engines to bolster conceptual understanding.
SPEC~\cite{peng2024synthesize} combines SAM~\cite{kirillov2023segment} and stable diffusion~\cite{rombach2022high} to generate fine-grained negative image samples.
However, these methods are hindered by the complexity of data generation and the potential for synthetic data to skew the consistency of the data distribution.

Our approach constructs negative image samples without relying on external models.
As shown in Figure~\ref{fig:comparison} (c), we capitalize on the VD embedded within our model to semantically modify input images in a novel end-to-end manner.

\section{Method}

This section details our \model~architecture, the NVA module, and the FGITM pretraining task.

\subsection{Model Architecture}
Given an image-text pair ($I$, $T$), the image $I$ is encoded into embeddings ${v_\mathrm{cls}, v_1,...,v_N}$, where $v_\mathrm{cls}$ is the \texttt{[CLS]} token's embedding, and $N$ denotes the number of image patches.
The text $T$ is similarly transformed into embeddings ${t_\mathrm{cls}, t_1,...,t_M}$, with $t_\mathrm{cls}$ corresponding to the text's \texttt{[CLS]} token embedding, and $M$ indicating the language encoder's maximum sequence length.
For visual features, all embeddings, except for the \texttt{[CLS]} token, are quantized into discrete tokens based on the VD and then concatenated with the \texttt{[CLS]} token to form an enhanced visual representation.
Our pretraining comprises two distinct stages. 
In the first stage, the quantized image embeddings are integrated with the encoded text embeddings through cross-attention mechanisms within the multi-modal encoder.
In the second stage, the quantized image embeddings are employed to acquire token-level negative image samples via our NVA module.
These samples, alongside positive image inputs, are fed into the multi-modal encoder.
The multi-modal encoder's output serves to pretrain and fine-tune downstream tasks.

\begin{figure*}[!htbp]
  \centering
  \includegraphics[width=\linewidth]{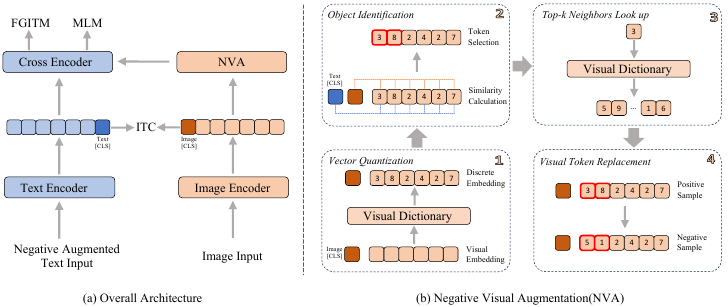}
  \caption{(a) The framework of the proposed end-to-end pretraining model NAS. (b) Illustration of our NVA. The continuous visual embedding encoded by the image encoder is firstly quantified into discrete embedding and then identifies the object in the image embedding based on the similarity between the global \texttt{[CLS]} embeddings and local discrete embeddings. We use the object embedding to search top-k neighbors in the dictionary and replace them with the neighbor tokens to construct negative image samples. [Best viewed in color.]}
  \label{fig:framework}
\end{figure*}

\subsection{Negative Visual Augmentation Module}

In this module, we introduce the VD as a fundamental component, which functions as a quantization framework to generate negative image samples. 
Formalized as a matrix $\mathcal{D} \in \mathbf{R}^{m \times c}$, it comprises $m$ vectors, each of dimension $c$. 
Initially randomized, the dictionary is progressively refined through a moving average process over mini-batches.
The process of associating each visual feature $v_i$ with an embedding vector in the dictionary $d_j$ is defined by:
\begin{equation}
h_i = \underset{d_j \in \mathcal{D}}{\text{argmin}}\quad\lVert v_{i}-d_{j}\rVert_{2},
\end{equation}
Updates to the VD within a mini-batch follow the equation:
\begin{equation}
\label{eqn:quant}
\hat{d_j} =\gamma * d_j + (1-\gamma) * \frac{\sum_{h_i=j}{v_i}}{n},
\end{equation}
with $\hat{d_j}$ representing the updated vector, $\gamma$ functions as the momentum coefficient (ranging from $[0,1]$), and $n$ is the count of visual patches mapped to $d_j$ within the current mini-batch—updating only when $n \neq 0$.
Since the $\text{argmin}$ operation is non-differentiable, we employ the stop-gradient operation to facilitate the visual encoder's training:
\begin{equation}
\hat{v}_i = \text{sg}[d_{h_i} - v_i] + v_i,
\end{equation}
where $\text{sg}[\cdot]$ denotes the stop-gradient operator, $h_i$ is an index in $\mathcal{D}$ and $v_i$ is subsequently assigned the value of $d_{h_i}$.

For the generation of negative image samples, we utilize the textual global feature $T_\mathrm{cls}$, the visual global feature $V_\mathrm{cls}$, and the visual local features $v_i$. 
Departing from conventional methods that rely solely on either $T_\mathrm{cls}$ or $V_\mathrm{cls}$ tokens for object identification~\cite{liang2022patches, jiang2022trips}, our approach synthesizes both to improve accuracy. 
We identify the primary object in an image by calculating a weighted sum $S$ of the cosine similarity $S_\mathrm{t}$ between $T_\mathrm{cls}$ and all local visual features $v_i$, and $S_\mathrm{v}$ between $V_\mathrm{cls}$ and $v_i$:
\begin{align}
S_\mathrm{t} &= cosine \langle T_\mathrm{cls}, v_i \rangle, \quad i = 1, \ldots, N \\
S_\mathrm{v} &= cosine \langle V_\mathrm{cls}, v_i \rangle, \quad i = 1, \ldots, N \\
S &= \lambda S_\mathrm{t} + (1-\lambda) S_\mathrm{v}
\end{align}
The hyper parameter ${\lambda}$ is used to control the weight of $S_\mathrm{t}$ and $S_\mathrm{v}$. 
In the experiment, we set it to 0.5.
The 30$\%$ tokens with the highest similarity score $S$ 
are identified as the primary subject of the image.
We then randomly replace these tokens with the $Top-k$ most similar tokens from the VD to construct token-level negative image samples. 
Specifically, we select the $Top-k$ embeddings in VD that are most similar to the current token(excluding itself), and randomly select one of them according to probability. 
Tokens with the same quantified index are replaced with the same embedding.
In the experiment, we set $k$ to 3.
Additionally, we incorporate 2-D sinusoidal positional embeddings using a sine function to enhance the model's spatial context comprehension. 
These negative image samples compel the encoder to recognize and encode subtle nuances, crucial for tasks requiring granular visual-textual discernment.

\subsection{Pretraining Tasks}
\noindent\textbf{Fine-Grained Image-Text Matching}
ITM predicts whether a given image-text pair is positive (matched) or negative (not matched), which is a binary classification task.
Based on ITM, the proposed FGITM aims to capture fine-grained differences of image-text pairs.
For each input image-text pair, we use two types of negative image samples: an in-batch negative image sample selected according to the similarity of images and texts in each mini-batch(the non-matching image with the highest similarity is selected as a negative image sample), and a token-level negative sample generated using the NVA module.
We use the multi-modal encoder’s output embedding of the \texttt{[CLS]} token as the joint representation of the image-text pair, and append a classification layer to predict the image-text matching probability $p^\mathrm{itm}$. The FGITM loss is a  cross-entropy loss: 
\begin{equation}
\label{eqn:itm}
\mathcal{L}_\mathrm{itm} = \mathbb{E}_{(I,T)\sim D} \mathcal{H} (y^\textrm{itm}, p^\textrm{itm}(I',T'))
\end{equation}
where $y^\textrm{itm}$ is a one-hot vector representing the ground-truth label. 
$(I',T')$includes $(I,T)$, $(I,T^{neg})$, $(I,T^{fg})$, $(I^{neg},T)$and $(I^{fg},T)$, where $I^{neg}/T^{neg}$ is the negtive image/text selected in every training batch and $I^{fg}/T^{fg}$ is the fine-grained negative image/text.

\noindent\textbf{Image-Text Contrastive Learning}
We follow the same settings of the ITC loss in ALBEF~\cite{albef}.
% ITC loss aims to learn the alignment of image and text representations.
Specifically, the similarity between image and text is calculated by the similarity function
$s(I,T) = l_v(v_\mathrm{cls})^\top l_t (t_\mathrm{cls})$,
where $l_v$ and $l_t$ are linear transformations that consist of a linear layer and a normalization layer. 
These transformations map $v_\mathrm{cls}$ and $t_\mathrm{cls}$ to normalized vectors in a reduced dimensional space.
Two queues are maintained to cache the most recently obtained $M$ image representations $I_m$ and $M$ text representations $T_m$, which are calculated by a momentum text encoder and a momentum image encoder respectively.
The normalized features obtained from the momentum model are denoted as
$l'_v (v'_\mathrm{cls})$ and $l'_t (t'_\mathrm{cls})$.
$s(I,T_m) = l_v(v_\mathrm{cls})^\top l'_t(t'_\mathrm{cls})$ and $s(T,I_m) = l_t(t_\mathrm{cls})^\top l'_v(v'_\mathrm{cls})$ 
define the similarity functions between the positive representations from the pretraining encoders and the negative representations from the momentum encoders. For each image and text, we compute the softmax-normalized image-to-text and text-to-image similarities as:
\begin{align}
\label{eqn:sim_i2t}
p^\mathrm{i2t}(I) &= \frac{\exp (s(I,T_m) / \tau)}{\sum_{m=1}^M \exp (s(I,T_m)/ \tau)}\\
\label{eqn:sim_t2i}
p^\mathrm{t2i}(T) &= \frac{\exp (s(T,I_m)/ \tau)}{\sum_{m=1}^M \exp (s(T,I_m)/ \tau)} 
\end{align}
where $\tau$ is a learnable temperature parameter.

Momentum distillation leverages the momentum model to distill the current training model, which is adapted to learn from pseudo-targets generated by the momentum model.
The final targets are:
\begin{align}
\label{eqn:tgt_i2t}
y^\mathrm{i2t}(I) &= (1-\alpha)y_\mathrm{one-hot}^\mathrm{i2t}(I)+\alpha p^\mathrm{i2t}(I_m)\\
\label{eqn:tgt_t2i}
y^\mathrm{t2i}(T) &= (1-\alpha)y_\mathrm{one-hot}^\mathrm{t2i}(T)+\alpha p^\mathrm{t2i}(T_m)
\end{align}
where $y_\mathrm{one-hot}^\mathrm{i2t}(I)$ and $y_\mathrm{one-hot}^\mathrm{t2i}(T)$ denote the ground-truth one-hot similarity.

The ITC loss over the pretraining dataset $D$ is defined as the cross-entropy $\mathcal{H}$ between $p$ and $y$:
\begin{equation}
\label{eqn:itc}
\mathcal{L}_\mathrm{itc} = \frac{1}{2} \mathbb{E}_{(I,T)\sim D} \big[ \mathcal{H}(y^\mathrm{i2t}(I),p^\mathrm{i2t}(I)) + \mathcal{H}(y^\mathrm{t2i}(T),p^\mathrm{t2i}(T)) \big]
\end{equation}

\noindent\textbf{Mask Language Modeling}
Masked Language Modeling(MLM) utilizes both the image and the contextual text to predict the masked words.
We randomly mask out the input tokens with a probability of 15\% and replace them with the special token \texttt{[MASK]} (following BERT, the replacements are 10\% random tokens, 10\% unchanged, and 80\% \texttt{[MASK]}).
Let $\hat{T}$ denotes a masked text, and $p^\textrm{msk}(I,\hat{T})$ denotes the predicted probability for a masked token. MLM minimizes a cross-entropy loss: 
\begin{equation}
\label{eqn:mlm}
\mathcal{L}_\mathrm{mlm} = \mathbb{E}_{(I,\hat{T})\sim D} \mathcal{H} (y^\textrm{msk}, p^\textrm{msk}(I,\hat{T}))
\end{equation}
where $y^\textrm{msk}$ is the one-hot vocabulary distribution.

The full pretraining objective of \model~ is:
\begin{equation}
\mathcal{L} = \mathcal{L}_\mathrm{itc}  + \mathcal{L}_\mathrm{mlm} + \mathcal{L}_\mathrm{itm}
\end{equation}	

\begin{table}[!t]
  \caption{Statistics of the pretraining datasets.}
  \label{tab:pretraining datasets}
  \centering
  \setlength{\tabcolsep}{1.7mm}
  \begin{tabular}{l  c  c  c  c  c}
    \toprule
    Dataset & MSCOCO & VG & SBU & CC-3M & \textbf{Sum.}\\
    \midrule
    \# images & 113K & 100K & 843K & 1.81M & 2.87M\\
    \# texts & 567K & 769K & 843K & 1.81M & 4.00M\\
    \bottomrule
  \end{tabular}
\end{table}

\begin{table*}[!t]
  \caption{Results on the ARO and Winoground benchmark. The NTA method yields substantial improvements on the ARO benchmark since it adopts task-specific hard negative types.}
  \label{tab:winoground}
  \centering
    \begin{tabular}{l  c  c  c  c  c   c  c  c  c}
    \toprule
    \multirow{2}{*}{Model} & \multirow{2}{*}{\#Images} 
    &\multicolumn{3}{c}{\textbf{ARO}} &\multicolumn{4}{c}{\textbf{Winoground}}\\
    & & Relation & Attribute &\textbf{Avg.} & Text & Image & Group & \textbf{Avg.}\\
    \midrule
    % MTurk Human &- & 89.50 &88.50 &85.50 &87.83\\
    Random Chance &-  &\multicolumn{3}{c}{50} & 25.0 & 25.0 & 16.7 & 22.2 \\
    \midrule
    % LXMERT &0.18M &19.25 &7.00 &4.00 &10.08\\
    % ViLBERT &3.1M &23.75 &7.25 &4.75 &11.92\\
    UNITER &4M &- &- &- &32.3 &13.3 &10.0 &18.5\\
    % ViLLA &4M &30.00 &12.00 &8.00 &16.67\\
    ViLT(ViT-B/32) &4M & 39.5 &20.3 & 29.9 &34.8 &14.0 &9.3 &19.3\\
    CLIP &400M &59.0 &62.0 &60.5 &30.8 &10.5 &8.0 &16.4\\
    FLAVA &60M &25.0 &73.0 &49.0 &25.3 &13.5 &9.0 &15.9\\
    % &ALBEF &4M &- &- &- &23.8 &13.8 &9.3 &15.6\\
    ALBEF$_\text{COCO}$ &4M &60.5 &88.5 &74.5 &27.5 &15.8 &11.0 &18.1\\
    \midrule
    \multicolumn{2}{l}{\textit{Large language models}}\\
    BART &- &81.1 &73.6 &77.4 &- &- &- &-\\
    FLAN-T5 &- &84.4 &76.5 &80.5 &- &- &- &-\\
    % &OPT &- &84.7 &79.8 &82.3 &- &- &- &-\\
    \midrule
    \multicolumn{2}{l}{\textit{Large Multi-modal models}}\\
    % &BEIT3 &- &60.6 &74.6 &67.6 &- &- &- &-\\
    BEIT3 &35M &60.6 &74.6 &67.6 &- &- &- &-\\
    LLaVA-7B &400M &- &- &- &13.5 &5.3 &2.3 &7.0\\
    MiniGPT-4 &500M &46.9 &55.7 &52.3 &23.3 &18.0 &9.5 &17.0\\
    \midrule
    \multicolumn{2}{l}{\textit{VD based models}}\\
    FDT &3M &49.8 &54.6 &52.2 &17.3 &3.5 &1.5 &7.4\\
    \midrule
    \multicolumn{2}{l}{\textit{Hard Negative based models}}\\
    NegCLIP &400M &80.2 &70.5 &75.4 &29.5 &10.5 &8.0 &16.0\\
    syn-CLIP &401M & 71.4 &66.9 &69.2 &30.0 &11.5 &9.5 &17.0\\
    SPEC &400M & 66.4 &73.7 &70.1 &- &- &- &-\\
    \midrule
    \textbf{NAS(NTA)$_\text{COCO}$} &2.9M &93.1 &91.7 & 92.4 &32.3 &17.3 &13.0 &20.8\\
    \textbf{NAS(NVA)$_\text{COCO}$} &2.9M &67.8 & 89.8 &78.8 &34.5 &19.0 &14.0 &22.5\\
    % \rowcolor{gray!25}
    \textbf{NAS(NTVA)$_\text{COCO}$} &2.9M &\textbf{93.2} &\textbf{93.4} &\textbf{93.3} &\textbf{35.3} &\textbf{22.0} &\textbf{18.5} &\textbf{25.3}\\
    \bottomrule
  \end{tabular}
\end{table*}

\begin{table*}[!t]
  \caption{Results on the VALSE benchmark.}
  \label{tab:VALSE}
  \centering
  \setlength{\tabcolsep}{1.5mm}
  \begin{tabular}	{l  c  c  c  c  c  c  c  c  c  c  c }
    \toprule	 	
    \multirow{2}{*}{Model} &\multirow{2}{*}{\#Images} &\textbf{Existence} &\textbf{Plurality}
    &\multirow{2}{*}{\textbf{Counting}} &\textbf{SP.rel.} &\multirow{2}{*}{\textbf{Action}} &\multirow{2}{*}{\textbf{Coreference}} &\multirow{2}{*}{\textbf{Foil-it!}} &\multirow{2}{*}{\textbf{Avg.}}\\
    & &quantifiers &number & &relations & & &  & \\
    \toprule
    LXMERT &0.18M &78.6 &64.4 &58.0 &60.2 &50.3 &45.5 &87.1 &63.5 \\
    ViLBERT &3.1M &65.5 &61.2 &65.1 &57.2 &69.5 &47.7 &86.9 &64.7 \\
    CLIP &400M & 66.9 &56.2 &60.7 &64.3 &72.1 &50.9 &88.8 &65.7 \\
    ALBEF$_\text{COCO}$ &2.9M & 75.4 &76.5 &65.8 &74.4 &67.5 &48.0 &92.6 &71.5\\
    XVLM$_\text{COCO}$ &4M & 83.0 &75.6 &67.5 &70.2 &71.2 &48.0 &94.8 &72.9\\
    FDT &3M &64.0 &56.8 &51.2 &51.8 &61.5 &47.3 &79.6 &58.9\\
    BLIP2 &500M &55.5 &71.5 &66.0 &62.4 &67.6 &50.3 &95.9 &67.0\\
    MiniGPT-4 &500M &65.5 &72.5 &67.4 &68.4 &71.0 &51.8 &95.8 &70.4\\
    \midrule
    \textbf{NAS(NTA)$_\text{COCO}$} &2.9M &85.5 &75.9  &66.8  &71.6 &75.5 &45.4 &93.7 &73.5\\
    \textbf{NAS(NVA)$_\text{COCO}$} &2.9M &85.1 &77.6  &66.7  &72.1 &72.7 &48.8 &94.2 &73.9\\
    % \rowcolor{gray!25}
    \textbf{NAS(NTVA)$_\text{COCO}$} &2.9M &87.3 &77.6  &70.1  &69.9 &75.8 &46.7 &93.2 &\textbf{74.4}\\
    \bottomrule
  \end{tabular}
\end{table*}

\section{Experiments}

\subsection{Pretraining Setup and Baselines}
\noindent\textbf{Pretraining Setup}
We use COCO~\cite{coco}, Visual Genome (VG)~\cite{vg}, Conceptual Captions (CC)~\cite{sharma2018conceptual}, and SBU Captions~\cite{ordonez2011im2text} as our pretraining datasets, which have a total of 4 million unique images and 5.1 million image-text pairs.
However, currently there are \textbf{only} 2.9 million available images and 4 million image-text pairs.
Detailed statistics are presented in Table \ref{tab:pretraining datasets}.
Our architecture leverages the initial six layers of BERT$_\mathrm{base}$ to initialize the text encoder, the subsequent six layers to initialize the multi-modal encoder, and DEiT-224/16 to initialize the image encoder.
The number of VD elements is set to 2,048.
In the NVA module, we set the balance parameter $\lambda$ to 0.5 and the parameter $k$ to 3.
For NTA and NTVA, to verify that our approach can work synergistically with existing methods, we fine-tune our model on the text-augmented
COCO dataset (Yuksekgonul et al. 2022).
Pretraining unfolds over 29 epochs in the first stage and a single epoch in the second stage, utilizing a batch size of 512.
We adopt the AdamW~\cite{loshchilov2019decoupled} optimizer with a weight decay of 0.02.
In the first 1000 iterations, the learning rate is warmed-up to $1e^{-4}$, and decayed to $1e^{-5}$ following a cosine schedule.
Each image is randomly cropped to $256\times256$ resolution, and RandAugment~\cite{randaugment} is adopted (we remove color changes from RandAugment because the text often contains color information).
During the fine-tuning stage, the resolution of an image is up-scaled to $384\times384$, and the positional encoding of the image patches is interpolated~\cite{vit}.
The momentum parameter for updating the momentum model is 0.995, and the queue length of cached features for image-text contrastive learning is set as 65, 536.
We linearly ramp-up the distillation weight $\alpha$ from 0 to 0.4 within the 1st epoch.
All experiments are performed on 8 NVIDIA A800 GPUs and take around 2 days to train. 

\begin{table*}[!t]
  \caption{Ablation study of VD and NVA.}
  \centering
  \label{tab:ablation module}
  \begin{tabular}{l  c  c  c  c  c  c  c  c  c  c}
    \toprule
    \multirow{2}{*}{Model} &\multirow{2}{*}{VD} &\multirow{2}{*}{NTA} &\multirow{2}{*}{NVA} &\multicolumn{3}{c}{Winoground} &\multicolumn{2}{c}{ARO} &\multirow{2}{*}{VALSE} &\multirow{2}{*}{\textbf{Avg.}}\\
    & & & &Text   &Image   &Group   &Relation   &Attribute &\\
    \midrule
    \textbf{NAS(wo/VD)} & & & &28.5 &15.0 &11.0 &59.2 &88.0 &71.5 &45.5\\
    \textbf{NAS(w/VD)} &\checkmark & & &34.8 &15.3 &13.8 &64.7 &88.7 &72.2 &48.3\\
    \textbf{NAS(NTA)} &\checkmark &\checkmark & &32.3 &17.3 &13.0 &93.1  &91.7 &73.5 &53.5\\
    \textbf{NAS(NTVA)} &\checkmark &\checkmark &\checkmark &\textbf{35.3} &\textbf{22.0} &\textbf{18.5} &\textbf{93.2}  &\textbf{93.4} &\textbf{74.4} &\textbf{56.1}\\
    \bottomrule
  \end{tabular}
\end{table*}

\noindent\textbf{Benchmarks} To test the effectiveness of our proposed NVA module, we have evaluated on three benchmarks.
\noindent{ARO}~\cite{negclip} is a large dataset designed for evaluating VLP models' object relational understanding and sensitivity to perturbations.
\noindent{Winoground}~\cite{thrush2022winoground} is a small dataset for evaluating compositional reasoning.
\noindent{VALSE}~\cite{Parcalabescu_2022} is designed for testing VLP models' visio-linguistic grounding capabilities. 

\noindent\textbf{Baselines} Our approach is benchmarked against several SOTA models. 
We primarily compare our model with multi-modal models and large language models.
For multi-modal models, we evaluate  LXMERT~\cite{tan2019lxmert}, ViLBERT~\cite{lu2019vilbert}, UNITER~\cite{chen2020uniter}, ViLT~\cite{vilt}, CLIP~\cite{clip}, ALBEF~\cite{albef}, XVLM~\cite{zeng2022multi}, FLAVA~\cite{singh2022flava}, NegCLIP~\cite{negclip}, syn-CLIP~\cite{cascante2023going}, SPEC~\cite{peng2024synthesize}, FDT~\cite{fdt}, BLIP2~\cite{blip2}, MiniGPT-4~\cite{zhu2023minigpt} and LLaVA~\cite{liu2024visual}. 
Among these, BLIP2, MiniGPT-4 and LLaVA are currently the most prominent multi-modal large language models. 
NegCLIP, syn-CLIP, and SPEC are based on hard negatives, while FDT is a VD-based model.
For large language models, we compare our model with BART~\cite{yuan2021bartscore} and FLAN-T5~\cite{chung2024scaling}.

\textbf{Detailed information about the benchmarks and baselines will be shown in Supplementary.}

\begin{figure}[!htbp]
  \centering
  \includegraphics[width=\linewidth]{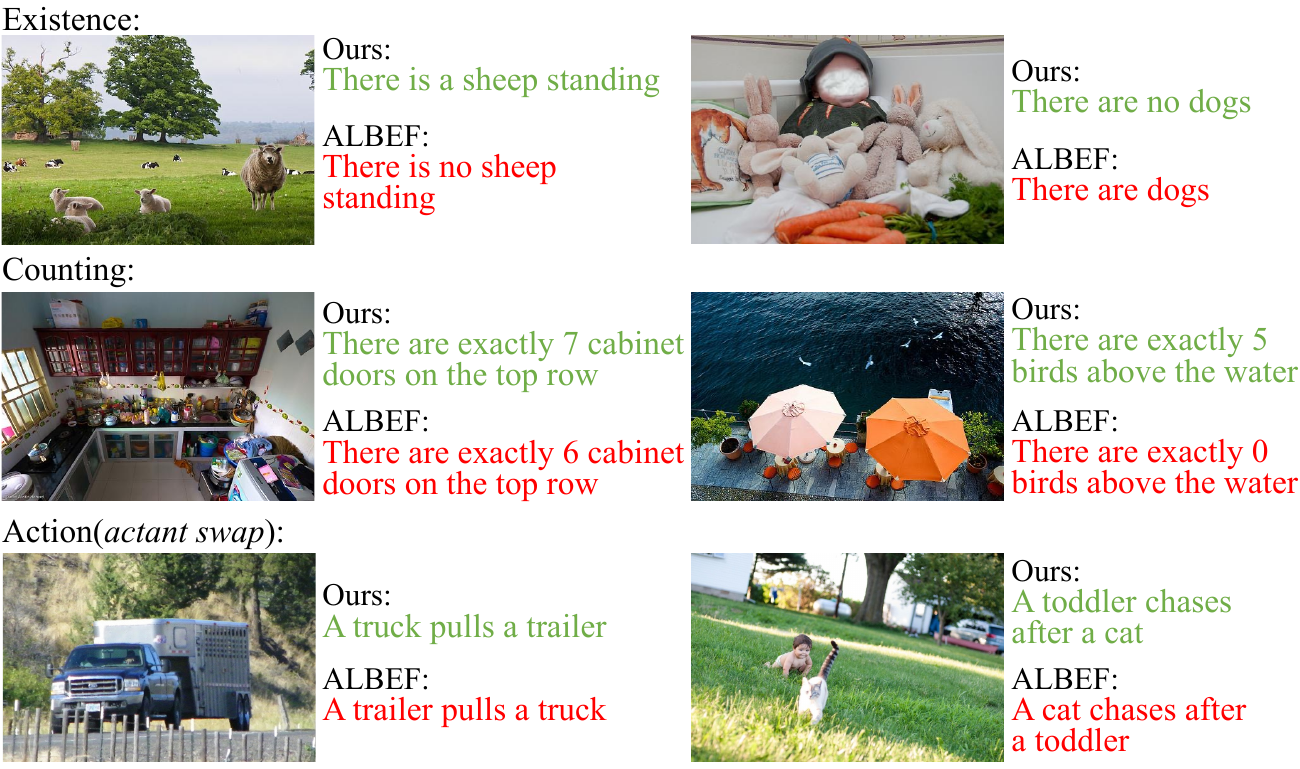}
  \caption{Cases on the VALSE benchmark. The first, second and third rows are the results on the existence, counting and actions(actant swap) test respectively. 
  More examples are seen in Supplementary.}
  \label{fig:cases}
\end{figure}

\subsection{Enhancement on Fine-grained Capability}
The evaluation includes results for the ARO (Table~\ref{tab:winoground}), Winoground (Table~\ref{tab:winoground}), and VALSE benchmark (Table~\ref{tab:VALSE}).
% The assessment criterion involves discerning between positive and negative captions associated with an image.
For NVA, we present fine-tuning results on the COCO dataset, which is denoted as \model$_\text{COCO}$.
For NTA and NTVA, we fine-tune our model on the text-augmented COCO dataset~\cite{negclip}.
For a fair comparison, we reproduce ALBEF on the training dataset we used and then fine-tuned on the COCO dataset.

Notably, even with less training data, \model~outperforms existing methods across these benchmarks (over the ALBEF baseline: an increase of 2.9\% on VALSE, 7.2\% on Winoground, and a remarkable 18.8\% on ARO). 
Such gains validate our approach's proficiency in leveraging hard negatives in both images and text.
For NVA, our method significantly improves the fine-grained capability, demonstrating the effectiveness of our approach.
In comparison, NTVA outperforms both NTA and NVA.
It proves that our NVA module can be compatible with the NTA methods.

The NTA in our model performs particularly well on the ARO benchmark, attributing its success to the dataset curated for specific hard text negatives.
Conversely, the NVA approach, through visual token replacement, ensures that our model attentively considers image details, thereby showing excellence in the VALSE and Winoground benchmark.
By introducing the NTVA method, the fine-grained feature extraction and fine-grained modality alignment of the model are significantly enhanced.
As depicted in Figure~\ref{fig:cases}, our model adeptly discerns subtle image nuances (for instance, discerning ``sheep standing'', ``no dogs'' and identifying ``5 birds''), showcasing refined fine-grained feature extraction skills. 
In conclusion, NTVA consistently delivers the most compelling results across all benchmarks, cementing the superiority of integrating NVA with NTA methodologies.
We also conducted a comparison on the broader retrieval task on COCO dataset. 
Despite utilizing less training data, our method demonstrates superior performance. 
Additionally, the use of exponential moving average mechanism to update VD results in the introduction of less than 1\% of the total parameters. 
To validate the generalizability of our method, we applied it to the BLIP model~\cite{blip}, which also led to improved performance. 
These results can be found in Supplementary.

\begin{table}[!t]
  \caption{Ablation study of the embedding vector size of VD.}
  \label{tab:ablation vd size}
  \centering
  \begin{tabular}{l  c  c  c  c  c}
    \toprule
    \multirow{2}{*}{VD size}  &\multicolumn{3}{c}{Winoground} &\multirow{2}{*}{VALSE} &\multirow{2}{*}{\textbf{Avg.}}\\
    &Text  &Image  &Group  & &\\
    \midrule
    \textbf{1024} &31.3 &\textbf{23.5} &14.8 &72.5 &35.3\\  
    \textbf{2048} &\textbf{35.3} &22.0 &\textbf{18.5} &\textbf{74.4} &\textbf{37.6}\\
    \textbf{4096} &29.8 &23.3 &15.3 &71.7 &35.0\\
    \textbf{8192} &27.8  &22.0 &12.8 &71.4 &33.5\\
    \bottomrule
  \end{tabular}
\end{table}

\begin{table}[!t]
  \caption{Ablation of the ratio of visual token replacement.}
  \label{tab:ablation replacement}
  \centering
  \begin{tabular}{l  c  c  c  c  c}
    \toprule
    \multirow{2}{*}{Ratio}  &\multicolumn{3}{c}{Winoground} &\multirow{2}{*}{VALSE} &\multirow{2}{*}{\textbf{Avg.}}\\
    &Text  &Image  &Group  & &\\
    \midrule
    \textbf{10\%} &30.3 &22.8 &16.5 &72.6 &35.6\\  
    \textbf{30\%} &\textbf{35.3} &22.0 &\textbf{18.5} &\textbf{74.4} &\textbf{37.6}\\
    \textbf{50\%} &34.0 &\textbf{23.8} &17.5 &73.4 &37.2\\
    \textbf{70\%} &32.5  &22.3 &16.3 &72.3 &35.9\\
    \bottomrule
  \end{tabular}
\end{table}

\begin{table}[!t]
  \caption{Ablation study of the balance parameter $\lambda$.}
  \label{tab:ablation lamda}
  \centering
  \begin{tabular}{c  c  c  c  c  c}
    \toprule
    \multirow{2}{*}{Value}  &\multicolumn{3}{c}{Winoground} &\multirow{2}{*}{VALSE} &\multirow{2}{*}{\textbf{Avg.}}\\
    &Text  &Image  &Group  & &\\
    \midrule
    \textbf{0.0} &30.3 &\textbf{24.0} &16.0 &73.2 &35.9\\ 
    \textbf{0.5} &\textbf{35.3} &22.0 &\textbf{18.5} &\textbf{74.4} &\textbf{37.6}\\
    \textbf{1.0} &33.0 &22.8 &15.0 &72.7 &35.9\\
    \bottomrule
  \end{tabular}
\end{table}

\subsection{Ablations}
We conducted ablation studies to assess the effectiveness of the NVA module and its synergy with the NTA methods. 
The effectiveness of VD and NVA is assessed in Table~\ref{tab:ablation module}. 
Additionally, we analyze the vector size $m$ of the VD in Table~\ref{tab:ablation vd size}.
Further, we conducted ablations on the replacement ratio of visual tokens, and the results are shown in Table~\ref{tab:ablation replacement}.
Lastly, the effects of the balance parameter ${\lambda}$ are detailed in Table~\ref{tab:ablation lamda}.

The introduction of the VD resulted in overall performance enhancements, which supports the role of the VD in enhancing fine-grained feature extraction. 
The integration of NTA further refined the alignment of fine-grained modality, leading to significant improvements. 
Finally, the introduction of NVA improved the model's performance, demonstrating the effectiveness of the proposed method and verifying the synergy between our method and existing NTA methods.
Our observations regarding the VD size revealed that a size of 2048 consistently achieved optimal results, aligning with the VD's design intent to consolidate similar visual semantics under unified image features. 
However, excessively granular semantic distinctions may hinder the extraction process of visual semantics and affect vision-language alignment. 
Conversely, a smaller VD size may impede fine-grained modality alignment. Empirically, a size of $m=2048$ yielded the most favorable outcomes and has thus been adopted as the default configuration. 

In the ablation experiment on the replacement ratio of visual tokens, a replacement ratio of 30\% yielded the best results in our experimental setting. 
We believe that a low visual token replacement ratio is not enough to constitute a negative visual sample, while a high replacement ratio will destroy the overall semantic information of the image and is not conducive to the convergence of the pretraining process.
Furthermore, in the ablation study of the balance parameter $\lambda$, setting $\lambda=0.5$ yielded optimal performance. This result suggests that simultaneously leveraging the global information from both the image and text is more effective for object identification within images.
Our results demonstrate the effectiveness of the NVA module, its synergy with existing NTA methods, and its versatile applicability across a wide range of tasks.

\section{Conclusions}

In this paper, we propose the NTVA method to simultaneously construct hard negative textual and visual samples. 
% Together with the FGITM task, \model~ is introduced to improve the fine-grained capability of VLP models.
The comprehensive experiments demonstrate the effectiveness of our \model~ and confirm that the NTVA method synergizes the hard negative samples, which greatly improves the fine-grained capability of \model, setting a new SOTA in the field. 
The NTVA method is a general data construction method that can be applied in related image fine-grained tasks.
In the future, we tend to integrate image segmentation approaches to recognize accurate and complete semantic regions for NVA. Meanwhile, we will investigate the co-quantize approach to align multi-modal information earlier and deeper.

\bibliography{aaai25}

\clearpage

\section{Supplementary for Benchmarks}

To test the effectiveness of our proposed Negative Visual Augmentation (NVA) module, we have evaluated on three benchmarks.
The statistics of benchmarks we use are shown in Table~\ref{tab:benchmark_data}.

\textbf{ARO}~\cite{negclip}
– or the Attribution, Relation, and Order benchmark, is a large dataset designed to evaluate the ability of Vision-Language Pretraining (VLP) models to understand four different types of skills.
It consists of Visual Genome Attribution and Visual Genome Relation, which leverages the Visual Genome~\cite{vg} dataset along with the GQA~\cite{hudson2019gqa} annotations to test the understanding of properties and relational understanding of objects in complex natural scenes.
VG-Relation includes 48 distinct relations with 23,937 test cases, and VG-Attribution includes 117 unique attribute pairs with 28,748 test cases.
It also leverages the COCO~\cite{coco} and Flickr30k~\cite{flickr30k} datasets to evaluate the model sensitivity to select the right caption after applying four different shuffling perturbations (e.g., exchanging nouns and adjectives, or by shuffling trigrams).
These tests are performed on the 5000 and the 1000 images from the respective COCO and Flickr30k test splits, respectively.

\textbf{Winoground}~\cite{thrush2022winoground}
– is a small dataset that evaluates the ability of VLP models for compositional reasoning, specifically understanding the meaning of the sentence after changing the order of its words.
The dataset has 400 samples, each comprised of two images and two texts.
The texts have the same words in a different order, each text corresponding to one image in the sample.
The Winoground metrics include (a) image score - percent of samples where the model picks the correct text for each image; (b) text score - percent of samples where the model picks the correct image for each text; (c) group score - percent of samples where both text and image score conditions are satisfied jointly.

\textbf{VALSE}~\cite{Parcalabescu_2022}
- is a novel benchmark designed for testing VLP models for their Vision-Language (VL) grounding capabilities on specific linguistic phenomena.
The dataset offers a suite of six tests covering a wide spectrum of basic linguistic phenomena affecting the linguistic and visual modalities: existence, plurality, counting, spatial relations, actions, and entity coreference.
The existence test challenges models to discern when entities are present or absent in images.
The plurality test checks a model's ability to distinguish singular from plural entities in images.
The counting test requires models to accurately count entities in an image. 
The relations test examines a model's understanding of spatial prepositions. 
The actions test evaluates if a model can match actions described in text to those depicted in images and identify the participants and their roles.
Finally, the coreference test checks if models can resolve pronominal references with visual grounding, relating pronouns in text to entities or regions in images.

\begin{table}
  \caption{Overview of the fine-grained VL benchmarks, where “IT” represents “Image-Text”.}
  \label{tab:benchmark_data}
  \centering
  \setlength{\tabcolsep}{1.5mm}
  \begin{tabular}	{l  c  c}
    \toprule	 	
    Benchmark &Task &\# IT pairs \\
    \midrule
    ARO-Relation &Relation &24k \\
    ARO-Attribution &Attribution &28.7k \\
    VALSE &Linguistic Phenomena &6.8k \\
    Winoground &Compositional Reasoning &1.6k \\
    \bottomrule
  \end{tabular}
\end{table}

\begin{table*}[!t]
  \caption{Zero-shot image-text retrieval results on Flickr30K and COCO.}
  \label{tab:retrieval_zeroshot}
  \centering
  \setlength{\tabcolsep}{1.9mm}
  \begin{tabular}	{l  c  c  c  c  c  c  c  c  c  c  c  c  c  c}
    \toprule	 	
    \multirow{3}{*}{Model} &\multirow{3}{*}{\# Images} & \multicolumn{6}{c}{Flickr30K (1K test set)} &\multicolumn{6}{c}{MSCOCO (5K test set)} \\
    &  &  \multicolumn{3}{c}{Text Retrieval}& \multicolumn{3}{c}{Image Retrieval} &  \multicolumn{3}{c}{Text Retrieval}& \multicolumn{3}{c}{Image Retrieval} \\
    & & R@1 & R@5 & R@10 & R@1 & R@5 & R@10 & R@1 & R@5 & R@10 & R@1 & R@5 & R@10 \\
    \midrule
    ImageBERT & 6M &70.7 &90.2 &94.0 &54.3 &79.6 &87.5 &44.0 &71.2 &80.4 &32.3 &59.0 &70.2\\
    UNITER$_\text{Large}$ & 4M &83.6 &95.7 &97.7 &68.7 &89.2 &93.9 &64.1 &87.7 &93.3 &48.8 &76.7 &85.8 \\
    ViLT & 4M &73.2 &93.6 &96.5 &55.0 &82.5 &89.8 & 56.5 & 82.6 & 89.6 & 40.4 & 70.0 & 81.1 \\
    CLIP & 400M & 88.0 &98.7 &99.4 &68.7 &90.6 &95.2 &58.4 &81.5 &88.1 &37.8 &62.4 &72.2 \\
    UNIMO$_\text{Base}$ & 4M &77.4 &95.1 &97.8 &62.4 &86.2 &91.7 &- &- &- &- &- &- \\
    ALBEF & 2.9M &88.1 & 98.3 & 99.3 &74.4 &92.0 &95.3 & 68.0 &89.5 &94.3 &49.2 &76.0 &84.6 \\
    UNIMO-2 & 19.8M &88.5 &96.8 &98.9 &72.7 &91.2 &94.6 &- &- &- &- &- &- \\
    \midrule
    \textbf{NAS(NVA)} & 2.9M &\textbf{90.8} & \textbf{98.9}& \textbf{99.7} &\textbf{76.8} &\textbf{93.2} &\textbf{96.4} & \textbf{70.9} & \textbf{90.6}  & \textbf{95.0}  & \textbf{52.7}& \textbf{79.0}  &\textbf{86.8}\\
    \bottomrule
  \end{tabular}
\end{table*}

\section{Supplementary for Baselines}

Our methodology is compared to various previous state-of-the-art (SOTA) models, summarized as follows.

\noindent\textbf{Multi-modal Models}

\begin{itemize}
    \item \textbf{LXMERT}~\cite{tan2019lxmert} learns the relationship between vision and language using Transformer~\cite{vaswani2017attention} encoders and a unique cross-modality encoder. It also enhances generalization through diverse pretraining tasks.
    \item \textbf{ViLBERT}~\cite{lu2019vilbert} extends the BERT~\cite{bert} architecture to a multi-modal two-stream model, processing both visual and textual inputs in separate streams that interact through co-attentional transformer layers.
    \item \textbf{UNITER}~\cite{chen2020uniter} introduces a conditional masking strategy on top of region-based image features to improve the fine-grained capability of the model.
    \item \textbf{ViLT}~\cite{vilt} assigns visual feature extraction to Transformer modules and avoids separate visual embedders, resulting in performance and efficiency improvements on VL tasks.
    \item \textbf{CLIP}~\cite{clip} exhibits robust zero-shot capabilities, trained via contrastive learning on large web-scale datasets, rivalling supervised methods in specific tasks.
    \item \textbf{ALBEF}~\cite{albef} first aligns the uni-modal image representation and text representation before fusing them with a multi-modal encoder, and implements momentum distillation to mitigate data noise.
    \item \textbf{XVLM}~\cite{zeng2022multi} performs “multi-grained vision language pretraining” to align texts with visual concepts in the images based on multi-granularity alignments.
    \item \textbf{FLAVA}~\cite{singh2022flava} excels across vision, language, and combined tasks with a tailored pretraining objective set.
    \item \textbf{Neg-CLIP}~\cite{negclip} presents composition-aware hard negative mining to generate textual hard negatives during model training to improve the compositional and order understanding of VLP models.
    \item \textbf{syn-CLIP}~\cite{cascante2023going} contributes a million-scale synthetic dataset and data generation pipeline using a 3D physics-based simulation platform to improve understanding and compositional reasoning of VLP models.
    \item \textbf{SPEC}~\cite{peng2024synthesize} introduces an efficient pipeline to synthesize candidate images that exclusively differ in a specific visual attribute and created the SPEC benchmark to diagnose the comprehension proficiency of VLP models.
    \item \textbf{FDT}~\cite{fdt} embeds both images and texts using a set of shared learnable discrete tokens, reducing the granularity gap between the two modalities.
    The matched visual and semantic concepts are enforced to be represented by the same set of discrete tokens by a sparse activation constraint.
    \item \textbf{BLIP2}~\cite{blip2} employs a lightweight Querying Transformer to bridge the modality gap. 
    Pretrained in two stages for VL representation and vision-to-language generative learning. 
    It demonstrates emerging capabilities in zero-shot instructed image-to-text generation.
    \item \textbf{MiniGPT-4}~\cite{zhu2023minigpt} aligns a frozen visual encoder with a frozen LLM, Vicuna, using just one projection layer.
    \item \textbf{LLaVA}~\cite{liu2024visual} uses language-only GPT-4 to generate multi-modal VL instruction-following data and connects a vision encoder and LLM for general purpose VL understanding.
\end{itemize}

\noindent\textbf{Large Language Models}

\begin{itemize}
    \item \textbf{BART}~\cite{yuan2021bartscore} is a generation-based baseline, using a pure LLM to compute text-only GPTScore.
    \item \textbf{FLAN-T5 }~\cite{chung2024scaling} excels in generalization performance and can perform well on more than 1800 NLP tasks.
\end{itemize}

\begin{table*}
  \caption{Fine-tuned image-text retrieval results on COCO.}
  \label{tab:retrieval_finetune_coco}
  \centering
  \begin{tabular}	{l  c  c  c  c  c  c  c}
    \toprule	 	
    \multirow{3}{*}{Model} &\multirow{3}{*}{\# Images} & \multicolumn{6}{c}{MSCOCO (5K test set)} \\
    &  &  \multicolumn{3}{c}{Text Retrieval}& \multicolumn{3}{c}{Image Retrieval} \\
    & & R@1 & R@5 & R@10 & R@1 & R@5 & R@10 \\
    \midrule
    ImageBERT & 6M &66.4 &89.8 &94.4 &50.5 &78.7 &87.1 \\
    UNITER$_\text{Large}$ & 4M &65.7 &88.6 &93.8 &52.9 &79.9 &88.0 \\
    ViLT & 4M & 61.5 & 86.3 & 92.7 & 42.7 & 72.9 & 83.1 \\
    SOHO & 200K & 66.4 & 88.2 & 93.8 & 50.6 & 78.0 & 86.7 \\
    ALBEF & 2.9M & 72.6 &91.5 &95.7 &55.2 &80.8 &\textbf{89.4} \\
    METER-Swin  & 4M & 73.0 &92.0 &96.3 &54.9 &81.4 &89.3 \\
    \midrule
    \textbf{NAS(NTA)} & 2.9M &62.7 &88.0  &94.0  &54.5 &80.4  &88.1\\
    \textbf{NAS(NVA)} & 2.9M & \textbf{74.0} & \textbf{92.7}  & \textbf{96.4}  &   \textbf{57.1}& \textbf{82.0}  &89.2\\
    \textbf{NAS(NTVA)} & 2.9M &63.1 &87.8  &94.2  &54.5 &80.2 &88.0\\
    \bottomrule
  \end{tabular}
\end{table*}

\section{Supplementary for Experiments}

This section provides an analysis of the performance of our model on the Image-Text Retrieval (ITR) task, which includes image-to-text retrieval (TR) and text-to-image retrieval (IR) subtasks. 
To assess the effectiveness of our pretrained model, we conducted experiments on the Flickr30K~\cite{flickr30k} and COCO~\cite{coco} datasets in both fine-tuning and zero-shot settings.
In the fine-tuning paradigm, the model, which was pretrained on a training set, underwent additional fine-tuning before evaluation on a separate validation/test set. 
Conversely, the zero-shot setting entailed evaluating the pretrained model directly on the test set. 
We benchmarked zero-shot retrieval on Flickr30K by employing the evaluation protocol established by ALBEF~\cite{albef}, which was fine-tuned on COCO.
For evaluation on ITR, we adopted a fine-tuning loss function comprising Image-Text Contrastive learning (ITC) and Image-Text Matching (ITM), with a re-ranking mechanism applied during inference. 
Initially, images and texts were encoded independently, and their respective similarity matrices were computed to yield the top-k candidates. 
Subsequently, candidate representations were input into a multi-modal encoder for re-ranking.

\begin{table*}
  \caption{Experiments on BLIP-based model exhibit similar performance trends.}
  \label{tab:blip_baseline}
  \centering
  \begin{tabular}	{l  c  c  c  c  c  c  c}
    \toprule	 	
    \multirow{2}{*}{BackBone} &\multirow{2}{*}{Methods} &\multicolumn{2}{c}{ARO} &\multicolumn{3}{c}{Winoground} &VALSE \\
    & &Rel. &Att. &Text &Image &Group &Avg. \\
    \midrule
    \multirow{2}{*}{ALBEF} &Fine-tune &60.5 &88.5 &27.5 &15.8 &11.0 &72.1 \\
    &\textbf{NTVA} &\textbf{93.2} &\textbf{93.4} &\textbf{35.3} &\textbf{22.0} &\textbf{18.5} &\textbf{74.4} \\
    \midrule
    \multirow{2}{*}{BLIP} &Fine-tune & 59.0 &88.0 &48.0 &23.8 &19.5 &73.3 \\
    &\textbf{NTVA} & \textbf{65.3} &\textbf{90.7} &\textbf{49.3} &\textbf{27.3} &\textbf{22.8} &\textbf{74.7} \\
    \bottomrule
  \end{tabular}
\end{table*}

We compare our model with ImageBERT~\cite{qi2020imagebert}, UNITER~\cite{chen2020uniter}, ViLT~\cite{vilt}, 
SOHO~\cite{soho}, CLIP~\cite{clip}, ALBEF~\cite{albef}, UNIMO~\cite{li2022unimo},  UNIMO-2~\cite{unimo2} and METER~\cite{dou2022empirical}. 
As depicted in Table~\ref{tab:retrieval_zeroshot} and Table~\ref{tab:retrieval_finetune_coco}, our method demonstrates superior performance compared with existing works on both zero-shot and fine-tuned retrieval benchmarks. 
Our NVA method excels against current VLP models, notably on the COCO dataset. 
Specifically, in the zero-shot scenario, our NAS (NVA) model surpasses ALBEF by 2.2\% in TR and 2.6\% in IR regarding R@1 on COCO.
For the fine-tuned retrieval, NAS (NVA) outperforms METER-Swin~\cite{dou2022empirical} by margins of 1.0\% for TR and 2.2\% for IR in terms of R@1 on COCO with less training data. 
It's important to highlight that while both Negative Textual Augmentation (NTA) and Negative Textual and Visual Augmentation (NTVA) methods showed underwhelming performance in retrieval tasks, the NTA method specifically caused the model to lose its capacity for capturing global information due to its focus on textual details. 
Our end-to-end NVA methodology does not suffer from this limitation, underscoring its advantages.

\noindent\textbf{Extending method on other backbones}
Our proposed method is model-agnostic and can be applied to other VLP models.
To verify it, we have extended our experiments to include BLIP~\cite{blip}. 
For a fair comparison, we compare our method with models fine-tuned on the COCO dataset.
The experimental results are presented in Table~\ref{tab:blip_baseline}.

Firstly, we observe that the BLIP-based model exhibits similar performance trends as the ALBEF-based model.
Specifically, our proposed method results in a significant enhancement across different benchmarks. 
This demonstrates that our proposed method can effectively enhance fine-grained capability and is transferable to a wide range of multi-modal models.
Moreover, by applying our proposed method to the superior base model BLIP, we achieve superior performance.

\section{Supplementary for Qualitative Examples}

This section presents a comprehensive qualitative evaluation of our model's performance on the VALSE and Winoground benchmarks. 

\begin{figure*}
  \centering
  \includegraphics[width=\linewidth]{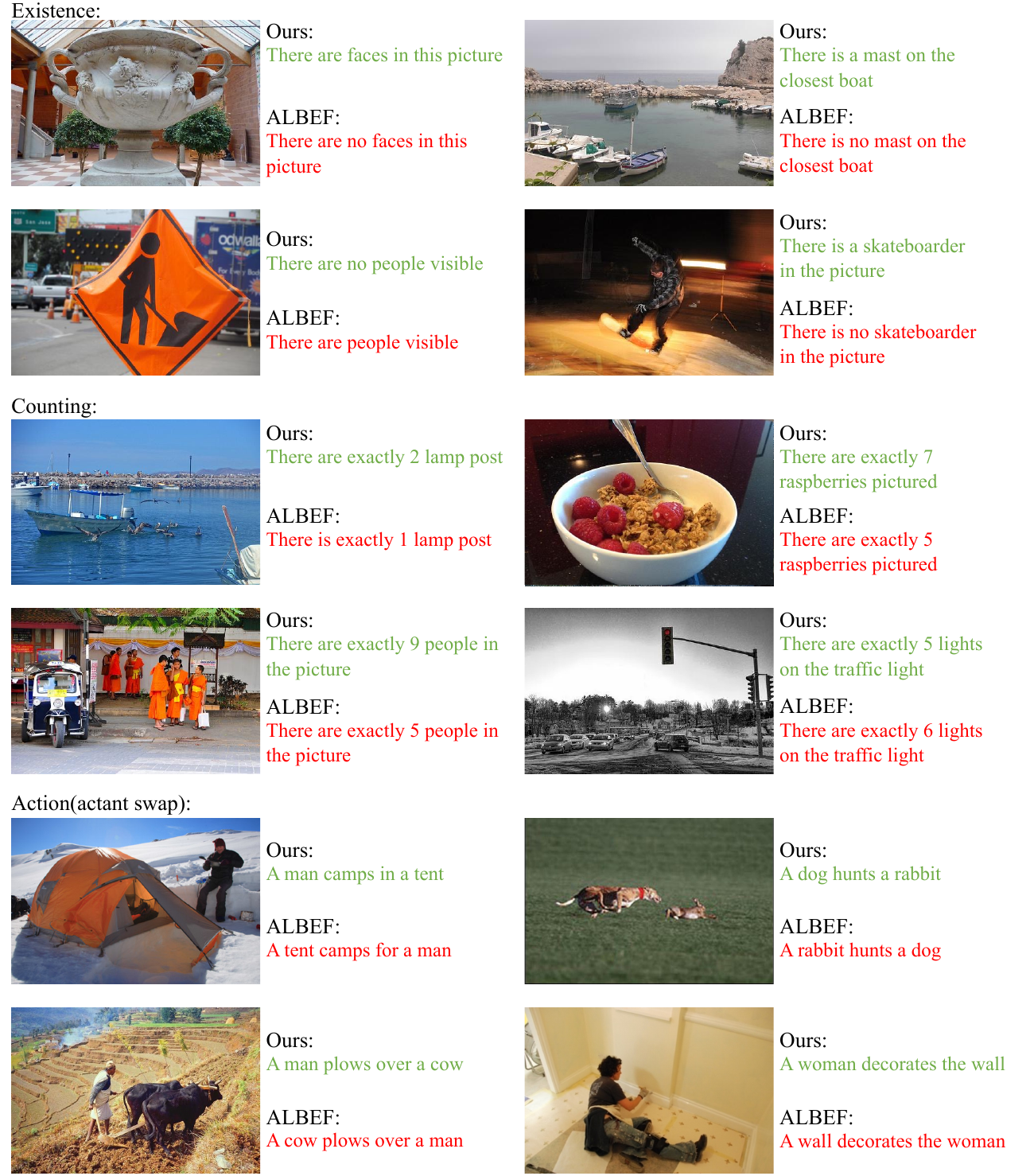}
  \caption{Examples on the VALSE benchmark. Sequentially from top to bottom, the panels display results for the existence test, counting test, and actions(actant swap) test respectively.}
  \label{fig:valse}
\end{figure*}

\begin{figure*}
  \centering
  \includegraphics[width=\linewidth]{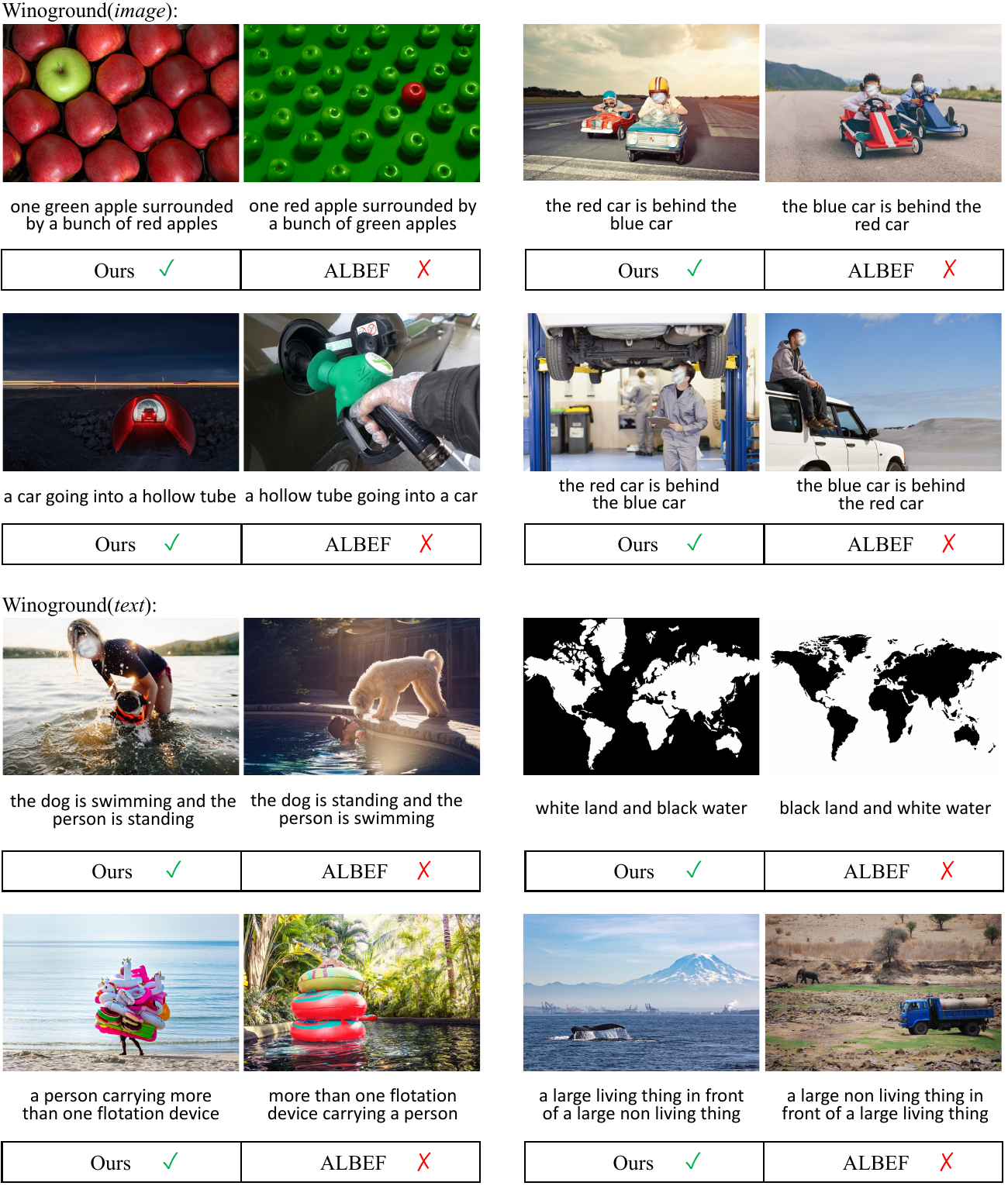}
  \caption{Examples on the Winoground benchmark. The first row illustrates the results of the image test, and the second row pertains to the text test.}
  \label{fig:winoground}
\end{figure*}

Figure~\ref{fig:valse} provides a selection of qualitative comparisons between our model and the baseline ALBEF model~\cite{albef} using the VALSE benchmark~\cite{Parcalabescu_2022}. 
These examples showcase our model's ability to understand abstract concepts, such as stylized facial features in artwork, and to identify subtle details, such as barely noticeable masts in images. 
Additionally, our model excels in tasks related to quantification and action understanding, demonstrating its sensitivity to subtle visual nuances and affirming the improvements in fine-grained feature extraction.
Figure~\ref{fig:winoground} offers additional insights where our model showcases enhanced performance over the baseline on the Winoground benchmark~\cite{thrush2022winoground}. 
The consistent success of our model in accurately pairing subjects with their attributes across diverse contexts indicates its advancements in fine-grained modality alignment and compositional reasoning. 

Overall, these examples illustrate the enhanced fine-grained capability of our model, confirming the effectiveness of our proposed method.

\end{document}